\pdfoutput=1

\documentclass[11pt]{article}

\usepackage[]{naacl2021}

\usepackage{times}
\usepackage{latexsym}
\usepackage{graphicx}
\usepackage{multicol}
\usepackage{multirow}
\usepackage{array}
\usepackage{enumitem}
\usepackage[framemethod=default]{mdframed}
\mdfdefinestyle{exampledefault}{%
rightline=true,innerleftmargin=10,innerrightmargin=10,
frametitlerule=true,frametitlerulecolor=black,
frametitlebackgroundcolor=lightgray,
frametitlerulewidth=2pt}
\usepackage{xcolor}
\newcolumntype{H}{>{\setbox0=\hbox\bgroup}c<{\egroup}@{}}

\usepackage[T1]{fontenc}

\usepackage[utf8]{inputenc}

\usepackage{microtype}

\title{Can Cross Encoders Produce Useful Sentence Embeddings?}

\author{Haritha Ananthakrishnan, Julian Dolby, Harsha Kokel, Horst Samulowitz, Kavitha Srinivas \\ hananthakris@ibm.com, dolby@us.ibm.com, harsha.kokel@ibm.com, \\ samulowitz@us.ibm.com, kavitha.srinivas@ibm.com
}

\begin{document}
\maketitle
\begin{abstract}
Cross encoders (CEs) are trained with sentence pairs to detect relatedness. As CEs require sentence pairs at inference, the prevailing view is that they can only be used as re-rankers in information retrieval pipelines.  Dual encoders (DEs) are instead used to embed sentences, where sentence pairs are encoded by two separate encoders with shared weights at training, and a loss function that ensures the pair's embeddings lie close in vector space if the sentences are related.  DEs however, require much larger datasets to train, and are less accurate than CEs.  We report a curious finding that embeddings from \textit{earlier layers} of CEs can in fact be used within an information retrieval pipeline.  We show how to exploit CEs to distill a lighterweight DE, with a 5.15x speedup in inference time.
\end{abstract}

\section{Introduction}
Following the introduction of the BERT model ~\cite{devlin2018bert}, encoder only retrieval models can be divided into two types: (a) dual encoder (DE)~\cite{lu2020twinbertdistillingknowledgetwinstructured}, where a model is trained by encoding each sentence in a pair by two encoders that share the same weights, and a loss function that ensures similar sentences have embeddings that are close in vector space, (b) cross encoder (CE) ~\cite{nogueira2020passagererankingbert, gao2021rethinktrainingbertrerankers} where a single encoder gets a sentence pair as input, and the model is trained to classify if the sentence pair is similar or not. Conventional wisdom is that DEs are less accurate than CEs~\cite{Hofstatter2020}. However, as CEs require a pair of sentences as input, they cannot be deployed in the initial retrieval phase due to the costs of computing the embeddings of the query with each document in the corpus.
A common information retrieval pipeline is therefore to use a DE for initial retrieval of the top K documents, whilst using CEs to rerank these top K results to improve accuracy. 

In this work, we report a curious empirical finding that the hidden states of earlier layers of CEs can be used for search.  This in itself should not be surprising: both the CE and DE were fine tuned from a pretrained model which has some representation of meanings of the sentences. However, we note that hidden states of earlier layers of the CE contain a good signal for sentence representations for IR, but later layers do not, presumably because later hidden states reflect the computation of differences between the pair of sentences. Additionally, hidden states of the earlier layers of the DE are \textit{significantly worse} than those of the CE for the same dataset, despite both sharing the same base pretrained model. The final layer of the DE is often better for accuracy of retrieval; the CE matches in some cases.  It appears that CEs extract information relevant for information retrieval in \textbf{\textit{earlier}} layers than DEs.  While DEs are commonly used as the base retrieval model because they can achieve low retrieval latency, training DEs is significantly more complex compared to CEs, often requiring knowledge distillation from cross encoder models to create a set of \textit{hard negatives}, as well as large training sets. We show how to exploit the signal in earlier layers of CEs to build more efficient sentence embedding models.
Our contributions are as follows:
\begin{itemize}
    \item We show that hidden states from earlier CE layers contain a stronger signal for information retrieval than earlier dual encoder layers.
    \item We build a 2 layer DE model using weights from earlier layers of a CE, as a new form of knowledge infusion.  This model is at least within 1\% accuracy of the baseline DE on 6/12 of the datasets.
    \item We show that inference is about 5 times faster on the 2-layer model.
\end{itemize}

\section{Related Work}
Analysis of BERT encoder attention was first performed by ~\cite{clark2019doesbertlookat}, but their focus was not knowledge distillation from CEs.

While DEs are commonly used for dense retrieval, recent work \cite{gao2021condenserpretrainingarchitecturedense} has proposed a better architecture called Condenser to build encoders.  Similarly, new techniques have been proposed for better hard negative mining to build better DEs~\cite{conf/iclr/XiongXLTLBAO21,DBLP:conf/naacl/QuDLLRZDWW21}.  Our focus is on knowledge distillation rather than changes in architectures or negative mining.   

DEs are much faster than CEs~\cite{humeau2019poly}, and CEs are more effective than DEs, so researchers have investigated approaches for cross-architecture distillation from a CE to a DE~\cite{Hofstatter2020,karpukhin2020densepassageretrievalopendomain,lu2022ernie,rauf2024bce4zsr}. Reranking is a common method for it where CEs are used after retrieval by either DEs or BM-25~\cite{YilmazYZL19,Nogueira2019,LogeswaranCLTDL19, gao2021rethinktrainingbertrerankers}. Shallow CEs have been proposed to use CEs to rerank documents more efficiently ~\cite{PetrovMM24}. Another example of CE use is a matrix factorization-based approach to efficiently approximate CE scores from a small subset of query-document pairs ~\cite{yadav2022efficient}. 
\section{Method}
\subsection{Embeddings extraction}
\begin{figure}
    \centering
    \includegraphics[width=.9\columnwidth]{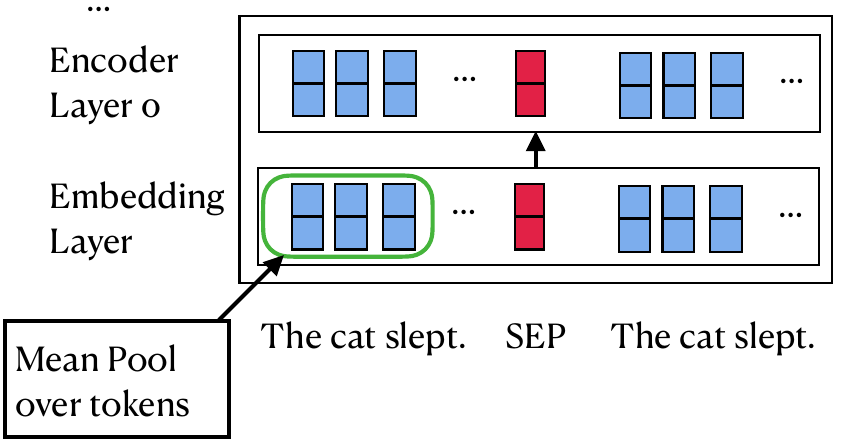}
    \caption{Extraction of layer-wise embeddings}
    \label{fig:embedding_extraction}
\end{figure}

Figure~\ref{fig:embedding_extraction} shows the approach to extract embeddings from a CE.  Since CEs are trained to process sentence pairs, to embed a single sentence, we simply pair it with itself.  To extract layer-wise embeddings, we mean pool over the tokens of the sentence for each layer, excluding padding and separator tokens.

\subsection{Knowledge infusion from cross encoder}
\label{ce-de}
\begin{figure}
    \centering
    \includegraphics[width=.9\columnwidth]{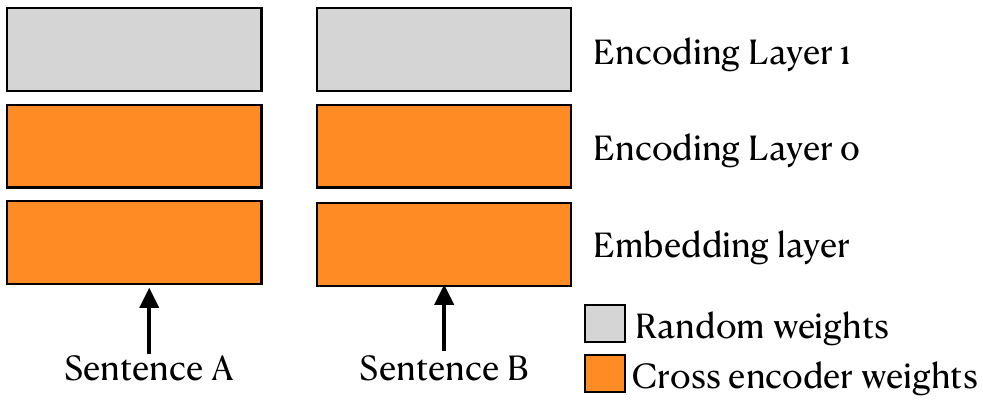}
    \caption{Knowledge infusion from cross encoder}
    \label{fig:dual_cross}
\end{figure}
Figure~\ref{fig:dual_cross} shows how to infuse the DE with structure learnt by the CE.  Weights from the embedding layer in BERT, which converts each input token to an input-vector embedding, and layer 0 from the BERT encoder are both copied into a DE model from the CE.  Encoding layer 1 was initialized to a set of random weights as usual, and the dual encoder was trained with the same ms-marco dataset,  with the training code from the sentence-transformer library \href{https://github.com/UKPLab/sentence-transformers/blob/master/examples/training/ms_marco/train_bi-encoder_mnrl.py}{(see here)}.  In using this training code, we used only BM-25 for hard negatives to avoid the need for knowledge distillation from additional models to create a DE from a CE. We term this model DE-2 CE in the rest of this paper.  Training time for this model was about 1 hour on an A100 GPU. We also compare DE-2 CE to a is a trained two-layered DE initialized with random initial weights. We call this model DE-2 Rand in this paper.

\subsection{Reranking}
\label{reranking}
Adhering to standard IR pipelines, in all our experiments we report retrieval metrics from the DE, and retrieve + rerank metrics obtained from re-ranking the top 50 most relevant results of the DE by a CE.
Reranking is done by the same CE used to distill information to the DEs mentioned in section \ref{ce-de}. We compare the final reranked results for the Baseline models chosen in \ref{models}, DE-2 CE, and  DE-2 Rand - as mentioned in \ref{ce-de} in Table \ref{tab:comp}.

\section{Experiments}
\subsection{Models}
\label{models}
We picked DE-CE model pairs with the same base pretrained model to compare. While massive benchmarks such as ~\cite{muennighoff2023mtebmassivetextembedding} are useful evaluating sentence encoders, we focused here on a smaller set of IR benchmarks just to demonstrate that a CE based knowledge infusion can be useful. As noted earlier, we made no attempt to ensure that both the DE and CE had the same training regimen since we used pretrained models; in fact, at least for the first pair of SBERT models, the training code to train the DE uses hard negatives from many different CEs, so that DE distils knowledge from multiple CEs. 
The model pairs used for comparison are:
\begin{itemize}[noitemsep,topsep=0pt]
    \item \href{https://www.sbert.net/docs/pretrained-models/msmarco-v3.html}{msmarco-MiniLM-L-12-v3} (dual encoder) and \href{https://huggingface.co/cross-encoder/ms-marco-MiniLM-L-12-v2}{cross-encoder/ms-marco-MiniLM-L-12-v2} (cross encoder)\footnote{We used v3 for the DE instead of v5 to keep the two versions as comparable as possible.}, each with 33M parameters.
    \item \href{https://huggingface.co/mixedbread-ai/mxbai-embed-large-v1} {mixedbread-ai/mxbai-embed-large-v1} (dual encoder) and \href{https://huggingface.co/mixedbread-ai/mxbai-rerank-large-v1}{mixedbread-ai/mxbai-rerank-large-v1} (cross encoder), with 335M and 435M parameters respectively.
\end{itemize}

\begin{figure}[!t]
    \centering
    \includegraphics[width=\linewidth]{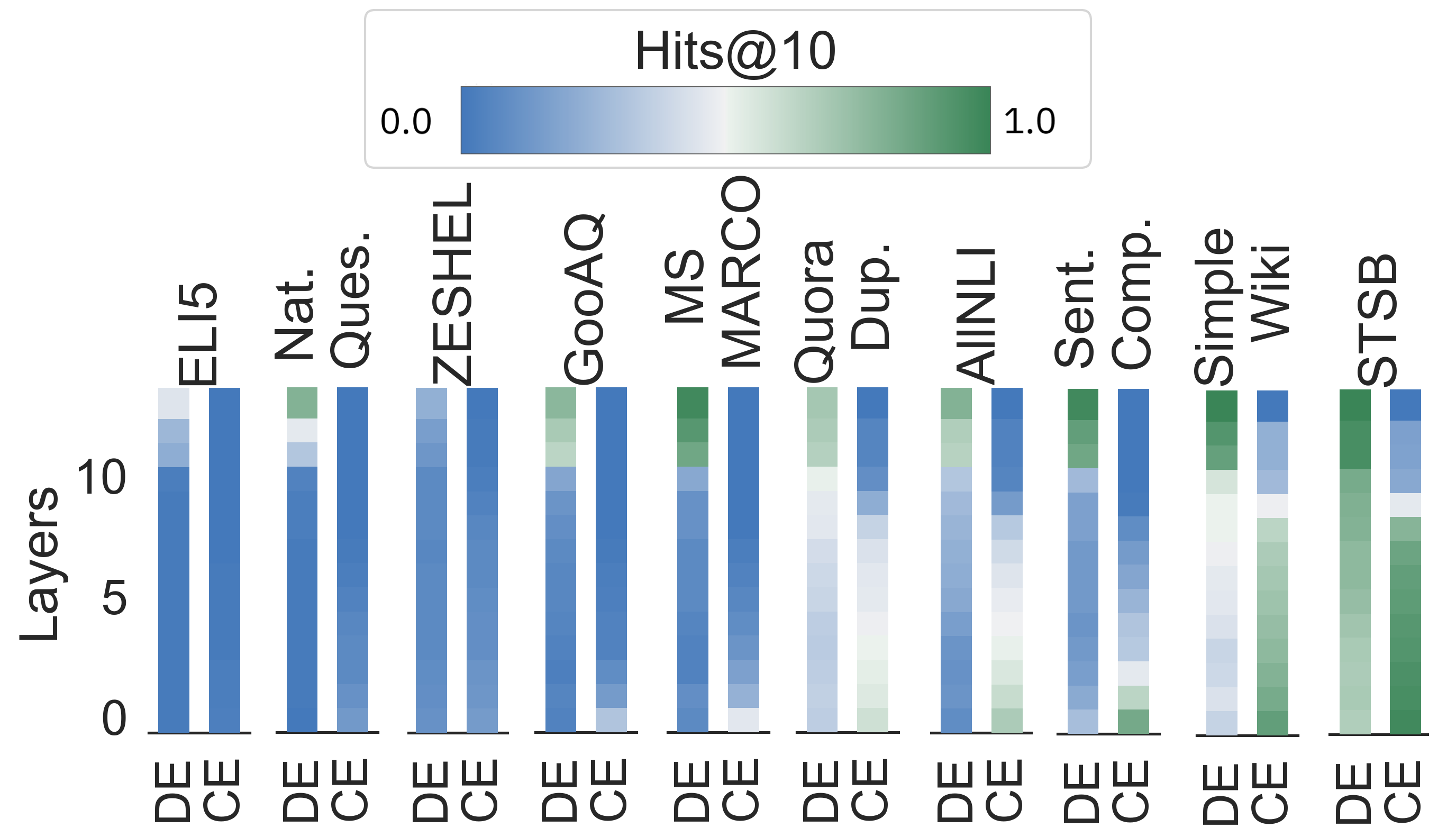}
    \caption{Comparison of layerwise performance of DE and CE on the msmarco pair of models. While performance improves monotonically with layers for DE, the CE embeddings from lower layers show surprisingly good performance, on many datasets.}
    \label{fig:msmarco_layer}
\end{figure}

\subsection{Datasets}
 Since embeddings are typically used for multiple purposes, we selected 12 datasets sampled to range over sentence similarity, question answering, and entity linking to cover a broader range of NLP tasks.\footnote{Task descriptions are available here: \href{https://sbert.net/docs/sentence_transformer/dataset_overview.html}{SBERT Datasets}.} We used datasets from the BEIR benchmark \cite{Thakur2021Beir} such as quora, fiqa, and scifact for testing the distilled DE model \cite{quora-question-pairs, Wadden2020Scifact, Maia2018Fiqa}. For ms-marco~\cite{nguyen2016ms}, we used the dev set, because the train split was used to train the first pair of models. For STSB, which scores sentence relatedness on a scale, we adapted it for a search task by using a 0.5 threshold to select sentence pairs. This dataset was chosen primarily because it was used in the original SBERT work~\cite{reimers-2019-sentence-bert}; zeshel was used because it was used in \cite{yadav2022efficient}. We also used the training data for other SBERT datasets such as all-nli, sentence-compression, natural-questions, eli5, and simplewiki~\cite{bowman-etal-2015-large, filippova-altun-2013-overcoming, 47761, fan-etal-2019-eli5, coster-kauchak-2011-simple}. Note that we only show layer by layer analysis for 10/12 datasets in Figures ~\ref{fig:msmarco_layer}, ~\ref{fig:msmarco}, and ~\ref{fig:mixed-bread}. This was done with a sample of 100,000 sentence pairs of the training data to show that these datasets exhibit a common pattern. However, the datasets shown in Table~\ref{tab:comp} are their respective full size training/dev datasets whose results are comparable to SOTA results.   
\begin{table*}[!th]
\centering
     \resizebox{1\textwidth}{!}{ \begin{tabular}{|l|llllll|l|lllll|}
\cline{1-7} \cline{9-13}
\textbf{}                  & \multicolumn{6}{c|}{\textbf{Retrieve}}                                                                                                                                                                                         & \textbf{} & \multicolumn{5}{c|}{\textbf{Retrieve + Rerank}}                                                                                                                                    \\ 
\cline{2-7} \cline{9-13}
{\textbf{dataset}}           & \multicolumn{3}{c|}{\textbf{Hits@10}}                                                                                   & \multicolumn{3}{c|}{\textbf{MRR@10}}                                                                & \textbf{} & \multicolumn{2}{c|}{\textbf{Hits@10}}                                         & \multicolumn{2}{c|}{\textbf{MRR@10}}                                           & \textbf{Speedup} \\

\cline{2-7} \cline{9-13} 
                           & \multicolumn{1}{l|}{\textbf{Baseline}} & \multicolumn{1}{l|}{\textbf{DE-2-Rand}} & \multicolumn{1}{l|}{\textbf{DE-2-CE}} & \multicolumn{1}{l|}{\textbf{Baseline}} & \multicolumn{1}{l|}{\textbf{DE-2-Rand}} & \textbf{DE-2-CE} & \textbf{} & \multicolumn{1}{l|}{\textbf{Baseline}} & \multicolumn{1}{l|}{\textbf{DE-2-CE}} & \multicolumn{1}{l|}{\textbf{Baseline}} & \multicolumn{1}{l|}{\textbf{DE-2-CE}} & \textbf{}        \\ 
                       \cline{1-7} \cline{9-13}
                       
msmarco/dev/small  & \multicolumn{1}{l|}{0.59}              & \multicolumn{1}{l|}{0.37}               & \multicolumn{1}{l|}{0.45}             & \multicolumn{1}{l|}{0.32}              & \multicolumn{1}{l|}{0.19}               & 0.24             &           & \multicolumn{1}{l|}{0.66}              & \multicolumn{1}{l|}{0.59}             & \multicolumn{1}{l|}{0.38}              & \multicolumn{1}{l|}{0.35}             & 5.1x             \\ \cline{1-7} \cline{9-13}
beir/quora/dev             & \multicolumn{1}{l|}{0.95}              & \multicolumn{1}{l|}{0.90}               & \multicolumn{1}{l|}{0.93}             & \multicolumn{1}{l|}{0.85}              & \multicolumn{1}{l|}{0.76}               & 0.81             &           & \multicolumn{1}{l|}{\textbf{0.96}}     & \multicolumn{1}{l|}{\textbf{0.96}}    & \multicolumn{1}{l|}{\textbf{0.74}}     & \multicolumn{1}{l|}{\textbf{0.74}}    & 5.3x             \\ \cline{1-7} \cline{9-13}
beir/scifact/test          & \multicolumn{1}{l|}{0.63}              & \multicolumn{1}{l|}{0.53}               & \multicolumn{1}{l|}{0.55}             & \multicolumn{1}{l|}{0.42}              & \multicolumn{1}{l|}{0.30}               & 0.36             &           & \multicolumn{1}{l|}{0.72}              & \multicolumn{1}{l|}{0.69}             & \multicolumn{1}{l|}{0.57}              & \multicolumn{1}{l|}{0.55}             & 4.7x             \\ \cline{1-7} \cline{9-13}
beir/fiqa/dev              & \multicolumn{1}{l|}{0.50}              & \multicolumn{1}{l|}{0.22}               & \multicolumn{1}{l|}{0.32}             & \multicolumn{1}{l|}{0.32}              & \multicolumn{1}{l|}{0.12}               & 0.18             &           & \multicolumn{1}{l|}{0.59}              & \multicolumn{1}{l|}{0.46}             & \multicolumn{1}{l|}{0.28}              & \multicolumn{1}{l|}{0.23}             & 5.3x             \\ \cline{1-7} \cline{9-13}
zeshel/test                & \multicolumn{1}{l|}{0.22}              & \multicolumn{1}{l|}{0.16}               & \multicolumn{1}{l|}{0.20}             & \multicolumn{1}{l|}{0.12}              & \multicolumn{1}{l|}{0.09}               & 0.11             &           & \multicolumn{1}{l|}{\textbf{0.20}}     & \multicolumn{1}{l|}{\textbf{0.19}}    & \multicolumn{1}{l|}{\textbf{0.11}}     & \multicolumn{1}{l|}{\textbf{0.10}}    & 4.8x             \\ \cline{1-7} \cline{9-13}
stsb/train                 & \multicolumn{1}{l|}{0.97}              & \multicolumn{1}{l|}{0.95}               & \multicolumn{1}{l|}{0.97}             & \multicolumn{1}{l|}{0.85}              & \multicolumn{1}{l|}{0.83}               & 0.85             &           & \multicolumn{1}{l|}{\textbf{0.98}}     & \multicolumn{1}{l|}{\textbf{0.98}}    & \multicolumn{1}{l|}{\textbf{0.88}}     & \multicolumn{1}{l|}{\textbf{0.88}}    & 4.7x             \\ \cline{1-7} \cline{9-13}
all-nli/train              & \multicolumn{1}{l|}{0.49}              & \multicolumn{1}{l|}{0.41}               & \multicolumn{1}{l|}{0.47}             & \multicolumn{1}{l|}{0.39}              & \multicolumn{1}{l|}{0.33}               & 0.36             &           & \multicolumn{1}{l|}{\textbf{0.55}}     & \multicolumn{1}{l|}{\textbf{0.54}}    & \multicolumn{1}{l|}{\textbf{0.47}}     & \multicolumn{1}{l|}{\textbf{0.46}}    & 4.5x             \\ \cline{1-7} \cline{9-13}
simplewiki/train           & \multicolumn{1}{l|}{0.97}              & \multicolumn{1}{l|}{0.98}               & \multicolumn{1}{l|}{0.98}             & \multicolumn{1}{l|}{0.92}              & \multicolumn{1}{l|}{0.93}               & 0.94             &           & \multicolumn{1}{l|}{\textbf{0.96}}     & \multicolumn{1}{l|}{\textbf{0.97}}    & \multicolumn{1}{l|}{\textbf{0.91}}     & \multicolumn{1}{l|}{\textbf{0.91}}    & 4.8x             \\ \cline{1-7} \cline{9-13}
natural-questions/train    & \multicolumn{1}{l|}{0.77}              & \multicolumn{1}{l|}{0.59}               & \multicolumn{1}{l|}{0.65}             & \multicolumn{1}{l|}{0.51}              & \multicolumn{1}{l|}{0.37}               & 0.42             &           & \multicolumn{1}{l|}{0.65}              & \multicolumn{1}{l|}{0.61}             & \multicolumn{1}{l|}{0.39}              & \multicolumn{1}{l|}{0.37}             & 5.4x             \\ \cline{1-7} \cline{9-13}
eli5/train                 & \multicolumn{1}{l|}{0.32}              & \multicolumn{1}{l|}{0.14}               & \multicolumn{1}{l|}{0.22}             & \multicolumn{1}{l|}{0.19}              & \multicolumn{1}{l|}{0.08}               & 0.12             &           & \multicolumn{1}{l|}{0.39}              & \multicolumn{1}{l|}{0.32}             & \multicolumn{1}{l|}{0.26}              & \multicolumn{1}{l|}{0.22}             & 5.4x             \\ \cline{1-7} \cline{9-13}
sentence compression/train & \multicolumn{1}{l|}{0.93}              & \multicolumn{1}{l|}{0.89}               & \multicolumn{1}{l|}{0.93}             & \multicolumn{1}{l|}{0.85}              & \multicolumn{1}{l|}{0.79}               & 0.84             &           & \multicolumn{1}{l|}{\textbf{0.96}}     & \multicolumn{1}{l|}{\textbf{0.96}}    & \multicolumn{1}{l|}{\textbf{0.93}}     & \multicolumn{1}{l|}{\textbf{0.94}}    & 6.6x             \\ \cline{1-7} \cline{9-13}
gooaq/train                & \multicolumn{1}{l|}{0.73}              & \multicolumn{1}{l|}{0.68}               & \multicolumn{1}{l|}{0.72}             & \multicolumn{1}{l|}{0.58}              & \multicolumn{1}{l|}{0.55}               & 0.58             &           & \multicolumn{1}{l|}{0.80}              & \multicolumn{1}{l|}{0.76}             & \multicolumn{1}{l|}{0.64}              & \multicolumn{1}{l|}{0.62}             & 5.2x             \\ \cline{1-7} \cline{9-13}
\end{tabular}
}
    \caption{Comparison of CE infused DE. \textbf{Baseline} is the pre-trained DE SBERT model.  \textbf{DE-2 Rand} is a trained two-layered DE initialized with random initial weights. \textbf{DE-2 CE} is a two-layered DE infused with initial weights from CE as explained in Fig~\ref{fig:dual_cross} and trained similar to DE-2. All the above models are trained on msmarco. Bold face numbers for DE-2 CE show where performance is at least within .01 of the baseline DE. \textbf{Speedup} is inference time gain for DE-2 CE over Baseline.}
    \label{tab:comp}
\end{table*} 
\subsection{Layer wise analysis}

\begin{figure}[!t]
    \centering
    \includegraphics[width=\linewidth]{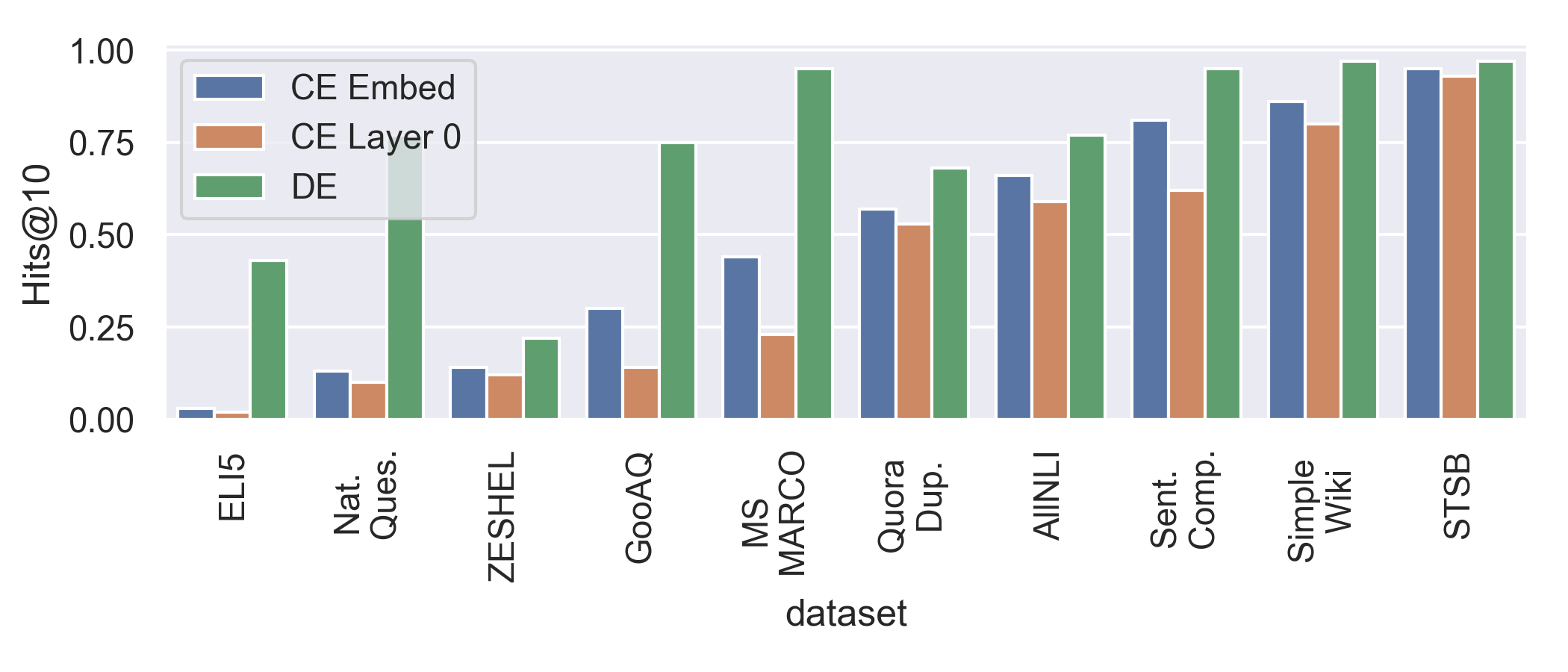}
    \caption{Performance of the DE final output layer (DE), CE embed, CE encoding layer 0 for the ms-marco pair of models.}
    \label{fig:msmarco}
\end{figure}

\begin{figure}[!t]
    \centering
    \includegraphics[width=\linewidth]{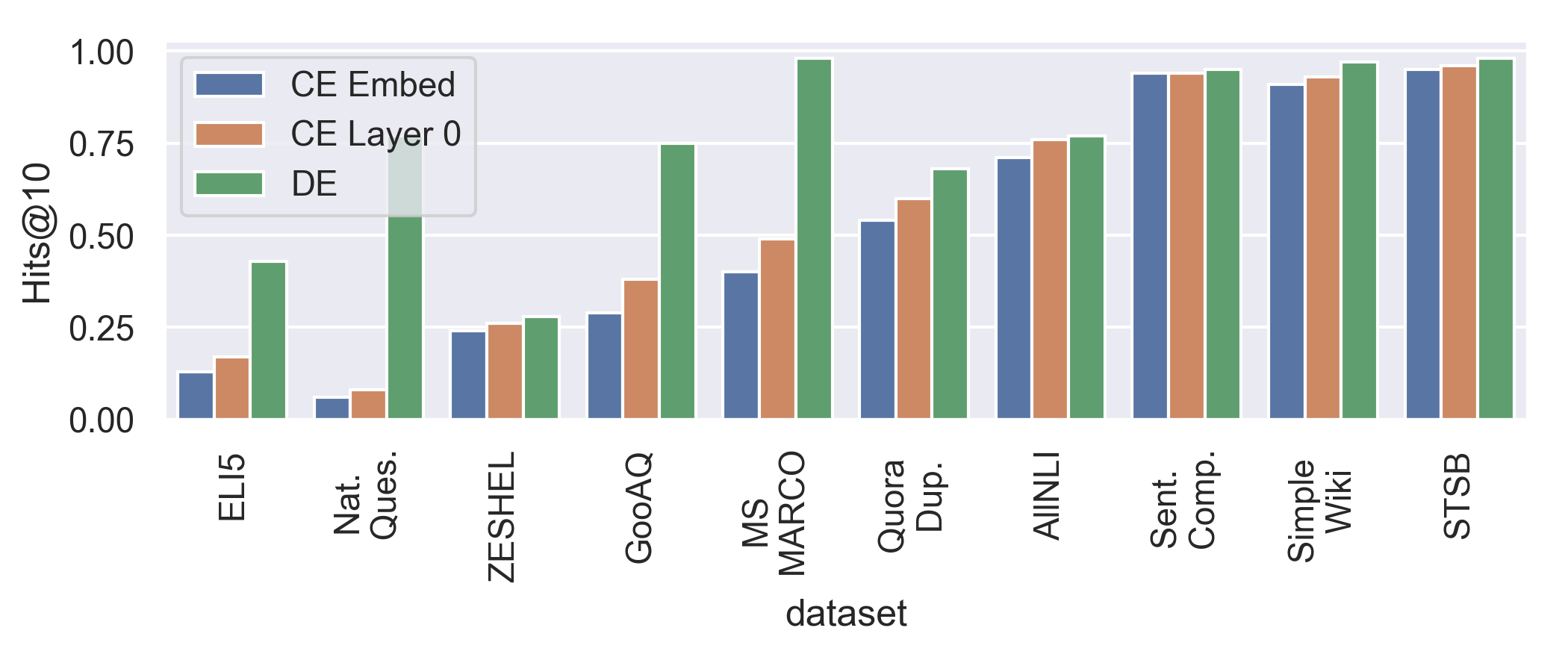}
    \caption{Performance of the DE final output layer (DE), CE embed, CE encoding layer 0 for the mixed bread pair of models.}
    \label{fig:mixed-bread}
\end{figure}
In this experiment, we report Top-10 cosine accuracy, or Hits@10\footnote{The evaluator used was    \href{https://sbert.net/docs/package_reference/sentence_transformer/evaluation.html}{Information Retrieval Evaluator}} 
as a metric for brevity; other information retrieval metrics showed a similar pattern. We also report MRR@10 in section~\ref{infused}.  As shown in Figure~\ref{fig:msmarco_layer}, CE hidden states in earlier layers contain a strong signal for information retrieval.  Surprisingly, hidden states from the DE's embedding layer were substantially worse than that of the CE by an average of 29.5\% across the 10 datasets for the ms-marco model pairs (paired two-tailed t test was significant, p < .01).  Further, Hits@10 for the CE's embedding layer was at 80\% or greater of the Hits@10 measure for the final output of the DE in 5/10 datasets; only in 2 cases was it under 30\%, see Figure~\ref{fig:msmarco}.

We observed the same pattern in the mxbai model pair\footnote{Layer wise analysis is not illustrated here due to space.}; Hits@10 for the DE embedding layer was worse than the CE embeddings by an average of 29.2\% (paired two-tailed t test was significant, p < .01).  Hits@10 performance of the CE was at 79\% or greater than Hits@10 performance of the final output of the DE in 5 of 10 datasets. Only in 1 case (natural ques.) was CE performance at below 30\% of the final DE performance, see Figure~\ref{fig:mixed-bread}. 


As shown in Figures~\ref{fig:msmarco} and~\ref{fig:mixed-bread}, CE embeddings are effective directly on some datasets (e.g. stsb and sentence-compression).

\subsection{Cross encoder infused dual encoders}
\label{infused}
Is it possible to leverage this CE knowledge in early layers to build more efficient DEs?  We trained the ms-marco models using the same training code from the \href{https://github.com/UKPLab/sentence-transformers/blob/master/examples/training/ms_marco/train_bi-encoder_mnrl.py}{sentence-transformers library}. Weights from the embedding layer and encoding layer 0 of the CE were copied to the DE as described in section~\ref{ce-de} to create a 15M parameter DE-2 CE\footnote{The step of how many layers of the CE need to copied to the DE was determined by experimentation, we tried copying 2 encoding layers or no encoding layers at all, just the embedding layer - this version had the best performance for this model.}. We compared it to a two layered DE with random initial weights to measure the impact of the CE in DE-2 CE and a DE-12 CE (a 12-layer Dual Encoder with CE initialized weights). Table~\ref{tab:comp} shows the results.  
\begin{itemize}[noitemsep,topsep=0pt]
    \item The average (Hits@10, MRR@10) for re-ranked DE-12 CE vs DE-2 CE was \textbf{(0.68, 0.55)} vs \textbf{(0.67, 0.54)}. This shows that the performance gain for  DE-12 CE is minimal even with 10 additional layers while offering no speedup advantage. Hence, this was omitted in Table ~\ref{tab:comp}.
    \item DE-2 CE is clearly better than DE-2 Rand.
    \item The difference between DE-2 CE after re-ranking whose weights were copied to the DE, was at least within 1\% of the 12 layered baseline DE with reranking on 6/12 datasets. We note that the baseline 12 layered DE was trained with hard negatives from multiple sources, so this result is significant. 
    \item On average DE-2 CE is only 0.99\% worse than the baseline DE across 12 datasets which was not statistically significant (paired two tailed t-test p > .19), but the \textbf{speedup for inference is on average 5.15x.}
\end{itemize}

\section{Conclusion}
We show that contrary to popular belief, 
CE can produce useful sentence embeddings from initial layers. Further, initial layers of CE 
can be used to build more efficient DEs with significant inference speedup. 
This has important implications for Retrieval Augmented Generation.  
\section{Limitations}

Our work covers limited sets of models and of datasets.  This is especially the case for the infusion of CE knowledge into DEs, since the fine-tuning code for mixed bread (mxbai) models was not available, unlike for msmarco.  

 There is a huge range of architectures for infusing a DE with CE knowledge.  We explored a small fraction of that space, and not exhaustively. 

 In the case of showing that the CE infused DE performs as well as the the baseline DE, we performed a t-test to show that the difference was not significant; but this is trying to prove the null hypothesis, which is not feasible.  We did show statistically significant better performance of earlier layers of the CE versus earlier layers of the DE, suggesting that the sample size was sufficient to detect those effect sizes.

 We explored a limited variety of NLP tasks, and there may be other use cases that benefit from our techniques to a greater or lesser extent.

 Finally, we note that all of our work was done on English text, and we did not study how it applies in other languages.

\section{Risks}
We don't see any risks other than the risks associated with using the original models.

\section{Ethical Concerns}

We cannot think of any ethical concerns that apply to this specific work beyond those that apply more generally to this domain.  

In fact, our work potentially makes dual encoders more efficient by reducing energy requirements.

\bibliography{anthology,custom}

\begin{thebibliography}{30}
\expandafter\ifx\csname natexlab\endcsname\relax\def\natexlab#1{#1}\fi

\bibitem[{Bowman et~al.(2015)Bowman, Angeli, Potts, and
  Manning}]{bowman-etal-2015-large}
Samuel~R. Bowman, Gabor Angeli, Christopher Potts, and Christopher~D. Manning.
  2015.
\newblock \href {https://doi.org/10.18653/v1/D15-1075} {A large annotated
  corpus for learning natural language inference}.
\newblock In \emph{Proceedings of the 2015 Conference on Empirical Methods in
  Natural Language Processing}, pages 632--642, Lisbon, Portugal. Association
  for Computational Linguistics.

\bibitem[{Clark et~al.(2019)Clark, Khandelwal, Levy, and
  Manning}]{clark2019doesbertlookat}
Kevin Clark, Urvashi Khandelwal, Omer Levy, and Christopher~D. Manning. 2019.
\newblock \href {http://arxiv.org/abs/1906.04341} {What does bert look at? an
  analysis of bert's attention}.

\bibitem[{Coster and Kauchak(2011)}]{coster-kauchak-2011-simple}
William Coster and David Kauchak. 2011.
\newblock \href {https://aclanthology.org/P11-2117} {{S}imple {E}nglish
  {W}ikipedia: A new text simplification task}.
\newblock In \emph{Proceedings of the 49th Annual Meeting of the Association
  for Computational Linguistics: Human Language Technologies}, pages 665--669,
  Portland, Oregon, USA. Association for Computational Linguistics.

\bibitem[{DataCanary et~al.(2017)DataCanary, hilfialkaff, Jiang, Risdal,
  Dandekar, and tomtung}]{quora-question-pairs}
DataCanary, hilfialkaff, Lili Jiang, Meg Risdal, Nikhil Dandekar, and tomtung.
  2017.
\newblock Quora question pairs.
\newblock \url{https://kaggle.com/competitions/quora-question-pairs}.
\newblock Kaggle.

\bibitem[{Devlin(2018)}]{devlin2018bert}
Jacob Devlin. 2018.
\newblock Bert: Pre-training of deep bidirectional transformers for language
  understanding.
\newblock \emph{arXiv preprint arXiv:1810.04805}.

\bibitem[{Fan et~al.(2019)Fan, Jernite, Perez, Grangier, Weston, and
  Auli}]{fan-etal-2019-eli5}
Angela Fan, Yacine Jernite, Ethan Perez, David Grangier, Jason Weston, and
  Michael Auli. 2019.
\newblock \href {https://doi.org/10.18653/v1/P19-1346} {{ELI}5: Long form
  question answering}.
\newblock In \emph{Proceedings of the 57th Annual Meeting of the Association
  for Computational Linguistics}, pages 3558--3567, Florence, Italy.
  Association for Computational Linguistics.

\bibitem[{Filippova and Altun(2013)}]{filippova-altun-2013-overcoming}
Katja Filippova and Yasemin Altun. 2013.
\newblock \href {https://aclanthology.org/D13-1155} {Overcoming the lack of
  parallel data in sentence compression}.
\newblock In \emph{Proceedings of the 2013 Conference on Empirical Methods in
  Natural Language Processing}, pages 1481--1491, Seattle, Washington, USA.
  Association for Computational Linguistics.

\bibitem[{Gao and Callan(2021)}]{gao2021condenserpretrainingarchitecturedense}
Luyu Gao and Jamie Callan. 2021.
\newblock \href {http://arxiv.org/abs/2104.08253} {Condenser: a pre-training
  architecture for dense retrieval}.

\bibitem[{Gao et~al.(2021)Gao, Dai, and
  Callan}]{gao2021rethinktrainingbertrerankers}
Luyu Gao, Zhuyun Dai, and Jamie Callan. 2021.
\newblock \href {http://arxiv.org/abs/2101.08751} {Rethink training of bert
  rerankers in multi-stage retrieval pipeline}.

\bibitem[{Hofst{\"{a}}tter et~al.(2020)Hofst{\"{a}}tter, Althammer,
  Schr{\"{o}}der, Sertkan, and Hanbury}]{Hofstatter2020}
Sebastian Hofst{\"{a}}tter, Sophia Althammer, Michael Schr{\"{o}}der, Mete
  Sertkan, and Allan Hanbury. 2020.
\newblock Improving efficient neural ranking models with cross-architecture
  knowledge distillation.
\newblock \emph{CoRR}, abs/2010.02666.

\bibitem[{Humeau et~al.(2019)Humeau, Shuster, Lachaux, and
  Weston}]{humeau2019poly}
Samuel Humeau, Kurt Shuster, Marie-Anne Lachaux, and Jason Weston. 2019.
\newblock Poly-encoders: Transformer architectures and pre-training strategies
  for fast and accurate multi-sentence scoring.
\newblock \emph{arXiv preprint arXiv:1905.01969}.

\bibitem[{Karpukhin et~al.(2020)Karpukhin, Oguz, Min, Lewis, Wu, Edunov, Chen,
  and Yih}]{karpukhin2020densepassageretrievalopendomain}
Vladimir Karpukhin, Barlas Oguz, Sewon Min, Patrick S.~H. Lewis, Ledell Wu,
  Sergey Edunov, Danqi Chen, and Wen{-}tau Yih. 2020.
\newblock Dense passage retrieval for open-domain question answering.
\newblock In \emph{{EMNLP} {(1)}}, pages 6769--6781. Association for
  Computational Linguistics.

\bibitem[{Kwiatkowski et~al.(2019)Kwiatkowski, Palomaki, Redfield, Collins,
  Parikh, Alberti, Epstein, Polosukhin, Kelcey, Devlin, Lee, Toutanova, Jones,
  Chang, Dai, Uszkoreit, Le, and Petrov}]{47761}
Tom Kwiatkowski, Jennimaria Palomaki, Olivia Redfield, Michael Collins, Ankur
  Parikh, Chris Alberti, Danielle Epstein, Illia Polosukhin, Matthew Kelcey,
  Jacob Devlin, Kenton Lee, Kristina~N. Toutanova, Llion Jones, Ming-Wei Chang,
  Andrew Dai, Jakob Uszkoreit, Quoc Le, and Slav Petrov. 2019.
\newblock Natural questions: a benchmark for question answering research.
\newblock \emph{Transactions of the Association of Computational Linguistics}.

\bibitem[{Logeswaran et~al.(2019)Logeswaran, Chang, Lee, Toutanova, Devlin, and
  Lee}]{LogeswaranCLTDL19}
Lajanugen Logeswaran, Ming{-}Wei Chang, Kenton Lee, Kristina Toutanova, Jacob
  Devlin, and Honglak Lee. 2019.
\newblock Zero-shot entity linking by reading entity descriptions.
\newblock In \emph{{ACL} {(1)}}, pages 3449--3460. Association for
  Computational Linguistics.

\bibitem[{Lu et~al.(2020)Lu, Jiao, and
  Zhang}]{lu2020twinbertdistillingknowledgetwinstructured}
Wenhao Lu, Jian Jiao, and Ruofei Zhang. 2020.
\newblock \href {http://arxiv.org/abs/2002.06275} {Twinbert: Distilling
  knowledge to twin-structured bert models for efficient retrieval}.

\bibitem[{Lu et~al.(2022)Lu, Liu, Liu, Shi, Huang, Sun, Tian, Wu, Wang, Yin
  et~al.}]{lu2022ernie}
Yuxiang Lu, Yiding Liu, Jiaxiang Liu, Yunsheng Shi, Zhengjie Huang, Shikun
  Feng~Yu Sun, Hao Tian, Hua Wu, Shuaiqiang Wang, Dawei Yin, et~al. 2022.
\newblock Ernie-search: Bridging cross-encoder with dual-encoder via self
  on-the-fly distillation for dense passage retrieval.
\newblock \emph{arXiv preprint arXiv:2205.09153}.

\bibitem[{Maia et~al.(2018)Maia, Handschuh, Freitas, Davis, McDermott, Zarrouk,
  and Balahur}]{Maia2018Fiqa}
Macedo Maia, S.~Handschuh, A.~Freitas, Brian Davis, R.~McDermott, M.~Zarrouk,
  and A.~Balahur. 2018.
\newblock Www'18 open challenge: Financial opinion mining and question
  answering.
\newblock \emph{Companion Proceedings of the The Web Conference 2018}.

\bibitem[{Muennighoff et~al.(2023)Muennighoff, Tazi, Magne, and
  Reimers}]{muennighoff2023mtebmassivetextembedding}
Niklas Muennighoff, Nouamane Tazi, Loïc Magne, and Nils Reimers. 2023.
\newblock \href {http://arxiv.org/abs/2210.07316} {Mteb: Massive text embedding
  benchmark}.

\bibitem[{Nguyen et~al.(2016)Nguyen, Rosenberg, Song, Gao, Tiwary, Majumder,
  and Deng}]{nguyen2016ms}
Tri Nguyen, Mir Rosenberg, Xia Song, Jianfeng Gao, Saurabh Tiwary, Rangan
  Majumder, and Li~Deng. 2016.
\newblock \href
  {https://www.microsoft.com/en-us/research/publication/ms-marco-human-generated-machine-reading-comprehension-dataset/}
  {Ms marco: A human generated machine reading comprehension dataset}.

\bibitem[{Nogueira and Cho(2020)}]{nogueira2020passagererankingbert}
Rodrigo Nogueira and Kyunghyun Cho. 2020.
\newblock \href {http://arxiv.org/abs/1901.04085} {Passage re-ranking with
  bert}.

\bibitem[{Nogueira and Cho(2019)}]{Nogueira2019}
Rodrigo~Frassetto Nogueira and Kyunghyun Cho. 2019.
\newblock Passage re-ranking with {BERT}.
\newblock \emph{CoRR}, abs/1901.04085.

\bibitem[{Petrov et~al.(2024)Petrov, MacAvaney, and Macdonald}]{PetrovMM24}
Aleksandr~V. Petrov, Sean MacAvaney, and Craig Macdonald. 2024.
\newblock Shallow cross-encoders for low-latency retrieval.
\newblock In \emph{{ECIR} {(3)}}, volume 14610 of \emph{Lecture Notes in
  Computer Science}, pages 151--166. Springer.

\bibitem[{Qu et~al.(2021)Qu, Ding, Liu, Liu, Ren, Zhao, Dong, Wu, and
  Wang}]{DBLP:conf/naacl/QuDLLRZDWW21}
Yingqi Qu, Yuchen Ding, Jing Liu, Kai Liu, Ruiyang Ren, Wayne~Xin Zhao, Daxiang
  Dong, Hua Wu, and Haifeng Wang. 2021.
\newblock \href {https://doi.org/10.18653/V1/2021.NAACL-MAIN.466} {Rocketqa: An
  optimized training approach to dense passage retrieval for open-domain
  question answering}.
\newblock In \emph{Proceedings of the 2021 Conference of the North American
  Chapter of the Association for Computational Linguistics: Human Language
  Technologies, {NAACL-HLT} 2021, Online, June 6-11, 2021}, pages 5835--5847.
  Association for Computational Linguistics.

\bibitem[{Rauf et~al.(2024)Rauf, Khalil, Wang, Wang, Ghani, and
  Hassan}]{rauf2024bce4zsr}
Muhammad~Arslan Rauf, Mian Muhammad~Yasir Khalil, Weidong Wang, Qingxian Wang,
  Muhammad Ahmad Nawaz~Ul Ghani, and Junaid Hassan. 2024.
\newblock Bce4zsr: Bi-encoder empowered by teacher cross-encoder for zero-shot
  cold-start news recommendation.
\newblock \emph{Information Processing \& Management}, 61(3):103686.

\bibitem[{Reimers and Gurevych(2019)}]{reimers-2019-sentence-bert}
Nils Reimers and Iryna Gurevych. 2019.
\newblock \href {https://arxiv.org/abs/1908.10084} {Sentence-bert: Sentence
  embeddings using siamese bert-networks}.
\newblock In \emph{Proceedings of the 2019 Conference on Empirical Methods in
  Natural Language Processing}. Association for Computational Linguistics.

\bibitem[{Thakur et~al.(2021)Thakur, Reimers, Rücklé, Srivastava, and
  Gurevych}]{Thakur2021Beir}
Nandan Thakur, Nils Reimers, Andreas Rücklé, Abhishek Srivastava, and Iryna
  Gurevych. 2021.
\newblock \href {https://arxiv.org/abs/2104.08663} {Beir: A heterogenous
  benchmark for zero-shot evaluation of information retrieval models}.
\newblock \emph{arXiv preprint arXiv:2104.08663}.

\bibitem[{Wadden et~al.(2020)Wadden, Lin, Lo, Wang, van Zuylen, Cohan, and
  Hajishirzi}]{Wadden2020Scifact}
David Wadden, Shanchuan Lin, Kyle Lo, Lucy~Lu Wang, Madeleine van Zuylen, Arman
  Cohan, and Hannaneh Hajishirzi. 2020.
\newblock \href {https://doi.org/10.18653/v1/2020.emnlp-main.609} {Fact or
  fiction: Verifying scientific claims}.
\newblock In \emph{Proceedings of the 2020 Conference on Empirical Methods in
  Natural Language Processing (EMNLP)}, pages 7534--7550, Online. Association
  for Computational Linguistics.

\bibitem[{Xiong et~al.(2021)Xiong, Xiong, Li, Tang, Liu, Bennett, Ahmed, and
  Overwijk}]{conf/iclr/XiongXLTLBAO21}
Lee Xiong, Chenyan Xiong, Ye~Li, Kwok{-}Fung Tang, Jialin Liu, Paul~N. Bennett,
  Junaid Ahmed, and Arnold Overwijk. 2021.
\newblock Approximate nearest neighbor negative contrastive learning for dense
  text retrieval.
\newblock In \emph{{ICLR}}. OpenReview.net.

\bibitem[{Yadav et~al.(2022)Yadav, Monath, Angell, Zaheer, and
  McCallum}]{yadav2022efficient}
Nishant Yadav, Nicholas Monath, Rico Angell, Manzil Zaheer, and Andrew
  McCallum. 2022.
\newblock Efficient nearest neighbor search for cross-encoder models using
  matrix factorization.
\newblock In \emph{{EMNLP}}, pages 2171--2194. Association for Computational
  Linguistics.

\bibitem[{Yilmaz et~al.(2019)Yilmaz, Yang, Zhang, and Lin}]{YilmazYZL19}
Zeynep~Akkalyoncu Yilmaz, Wei Yang, Haotian Zhang, and Jimmy Lin. 2019.
\newblock Cross-domain modeling of sentence-level evidence for document
  retrieval.
\newblock In \emph{{EMNLP/IJCNLP} {(1)}}, pages 3488--3494. Association for
  Computational Linguistics.

\end{thebibliography}
\bibliographystyle{acl_natbib}

\end{document}